\documentclass{article}

\usepackage{arxiv}

\usepackage[utf8]{inputenc} % allow utf-8 input
\usepackage[T1]{fontenc}    % use 8-bit T1 fonts
\usepackage{hyperref}       % hyperlinks
\usepackage{url}            % simple URL typesetting
\usepackage{booktabs}       % professional-quality tables
\usepackage{amsfonts}       % blackboard math symbols
\usepackage{nicefrac}       % compact symbols for 1/2, etc.
\usepackage{microtype}      % microtypography
\usepackage{lipsum}
\usepackage{graphicx}
\graphicspath{ {./images/} }

\title{TFR: Texture Defect Detection with Fourier Transform using Normal Reconstructed Template of Simple Autoencoder
}

\author{
 Jongwook Si \\
  Dept. Computer AI Convergence Engineering\\
  Kumoh National Institute of Technology\\
  Gumi, KOREA 39177 \\
  \texttt{jwsi425@kumoh.ac.kr} \\
   \And
 Sungyoung Kim \\
  Dept. Computer Engineering\\
  Kumoh National Institute of Technology\\
  Gumi, KOREA 39177 \\
  \texttt{sykim@kumoh.ac.kr} \\
  \And
}

\begin{document}
\maketitle
\begin{abstract}
Texture is an essential information in image representation, capturing patterns and structures. As a result, texture plays a crucial role in the manufacturing industry and is extensively studied in the fields of computer vision and pattern recognition. However, real-world textures are susceptible to defects, which can degrade image quality and cause various issues. Therefore, there is a need for accurate and effective methods to detect texture defects. In this study, a simple autoencoder and Fourier transform are employed for texture defect detection. The proposed method combines Fourier transform analysis with the reconstruct-ed template obtained from the simple autoencoder. Fourier transform is a powerful tool for analyzing the frequency domain of images and signals. Moreover, since texture defects often exhibit characteristic changes in specific frequency ranges, analyzing the frequency domain enables effective defect detection. The proposed method demonstrates effectiveness and accuracy in detecting texture defects. Experimental results are presented to evaluate its performance and compare it with existing approaches.
\end{abstract}

% keywords can be removed
\keywords{ Texture\and Defect Detection\and Anomaly Detection\and Fourier Transform\and Reconstruction}

\section{Introduction}
Texture defect detection is a crucial topic in image processing and computer vision, as texture provides important information about patterns and structures in images. Texture defects are particularly significant in the manufacturing industry, where they are essential for evaluating product quality and detecting defects in manufacturing processes.\\
In the manufacturing industry, it is crucial to ensure that the produced products meet the required standards. Texture defects can significantly impact product quality. Texture defect detection allows for the identification and resolution of anomalies or irregularities occurring in texture pat-terns or structures, enabling the evaluation of product quality. The paper focuses on proposing a new approach that combines Fourier transform and autoencoder-based methods to effectively detect and address texture defects.\\
Texture defects can arise during the manufacturing process, and detecting and identifying such defects is essential to maintain accuracy. By applying texture defect detection techniques, manufacturing companies can swiftly detect issues and take prompt actions. Furthermore, texture defect detection offers advantages in terms of cost reduction. Texture defects can lead to additional costs for manufacturing companies, such as customer dissatisfaction. By implementing robust texture defect detection methods, companies can minimize the occurrence of defective products, reduce waste, and optimize production efficiency.\\
Manual inspection of texture defects is time-consuming and subjective, posing limitations. The automation of the texture defect detection process using computer vision technology enables faster and more objective evaluations. The paper aims to contribute to the improvement of product quality, optimization of manufacturing processes, cost reduction, and automation by advocating the importance of texture defect detection and proposing a novel approach that integrates Fourier trans-form and autoencoder-based methods.\\
Various techniques have been used in texture defect detection, including the combination of deep learning and image processing methods. In this study, Fourier transform and a simple autoencoder are utilized for texture defect detection. Fourier transform is a useful tool for analyzing the frequency domain of images or signals, enabling the characterization of texture frequency proper-ties. Since texture defects may exhibit characteristic changes in certain frequency ranges, analyzing the frequency domain using Fourier transform allows for effective defect detection. Additionally, the autoencoder is employed to learn the normal characteristics of textures and reconstruct them. An autoencoder is a model that encodes input data into a latent space and reconstructs it, thereby extracting features from the input data. In this research, the autoencoder is used to reconstruct nor-mal texture data and generate a template of normal reconstruction. This template is utilized for de-fect detection by analyzing the differences between the reconstructed texture and the normal state. This paper provides a detailed explanation of the reconstruction process of the autoencoder and the method for generating the normal reconstruction template for texture defect detection. The aim is to contribute a fresh perspective and advancements to the field of texture defect detection. \\

The contributions of this paper are as follows:\\

•	Performance Improvement: The proposed simple autoencoder architecture achieves a high level of performance. Despite its simplicity, the autoencoder demonstrates effectiveness in texture defect detection, proving that efficient defect detection can be achieved without complex deep learning models.\\
•	Integration of Techniques: This study combines deep learning and image processing techniques for texture defect detection. Deep learning is used for denoising and reconstruction, while image processing methods are employed for extracting texture features and defect detection. This combination offers advantages such as not requiring extensive data training. Thus, the research presents a concise and effective approach for texture defect detection by integrating deep learning and image processing techniques.\\
•	Experimentation and Analysis: Detailed experiments and analyses are conducted on the necessary parameters. Various parameters are adjusted and compared to optimize the performance of texture defect detection. This provides insights into the parameters that have the most significant impact on performance and offers practical guidelines for real-world applications.\\

Through these contributions, this paper demonstrates the potential for performance enhance-ment in texture defect detection and the benefits of combining deep learning and image processing techniques.

\section{Related Works}
\label{sec:headings}
In anomaly detection research, methods based on reconstruction are widely studied. These methods typically involve training on normal data to generate reconstructed data and utilize the difference between the input anomalous data and the original image for detection. Models such as AutoEncoder and GAN are commonly employed for reconstruction and leveraging reconstruction errors for anomaly detection.

\subsection{Anomaly Detection with Reconstruction}

AnoGAN[1] proposes a basic approach to anomaly detection by combining unsupervised learning with GAN. It learns the distribution of normal data by inputting only normal data and calculates anomaly scores to compare and detect anomalies. f-AnoGAN[2], an extension of AnoGAN[1], improves performance by introducing fast mapping techniques for new data and incorporating an Encoder into GAN for more refined reconstruction. The generated data by f-AnoGAN[2] exhibits high generation performance that even experts find it difficult to distinguish from real data. GANomaly[3] is a method that learns both generation and latent space by using only normal data. Anomaly scores are computed based on the differences in latent vectors. Skip-GANomaly[4] ex-tends GANomaly[3] with a U-Net-based network architecture and introduces adversarial training that includes a loss function for Discriminator's feature maps, leading to improved reconstruction performance. MemAE [5] improves the limitations of using AutoEncoder for anomaly detection through the incorporation of a Memory Module, which makes reconstruction more challenging for abnormal samples. While AutoEncoder generalizes well, it can also reconstruct abnormal regions, which is a drawback. This paper focuses on post-processing methods to address this issue. OCGAN [6] is a model for one-class anomaly detection, where it learns latent representations of in-class examples and restricts the latent space to the given class. By utilizing a denoising AutoEncoder net-work and a discriminator, it generates in-class samples and explores anomalies outside the class boundaries. This approach achieves high-performance results.\\
These studies[1-6] mainly focus on reconstruction-based methods for anomaly detection, where training on normal data is used to assess and detect anomalies. Anomaly detection encompasses various subfields[7-9] that can be applied in real-life scenarios, including disease detection, accident detection, and fall detection, among others.\\
Y. Zhao et.al.[7]’s the objective is to effectively detect diseases in plants. However, the paper mentions the issue of data imbalance between diseased and healthy samples and proposes a solution called DoublaGAN, which applies Super-Resolution to augment the diseased data. This re-search achieves the goal of detecting various plant diseases while enhancing the resolution by a factor of four and demonstrates high performance. J. Si et.al.[8] introduces the structure of a Generative Adversarial Serial Autoencoder, which consists of Autoencoders connected in series. The pa-per presents a method for detecting diseases in chili peppers by using the Grabcut technique to segment the pepper regions and applying a reconstruction process. It addresses the limitations of reconstruction and improves performance by calculating all scores based on reconstruction. J. Si et.al.[9] is focused on detecting traffic accidents in black box videos. The proposed method generates the next frame of the video using information from the previous frames and compares it with the actual frame to detect accidents. However, the paper mentions the need to address the issue of misclassification in the background when dealing with moving videos.

\subsection{Defect Detection}

D. M. Tsai et.al.[10], a study proposing the use of Fourier transform to detect defects in PCBs is presented. The research demonstrates the ability to detect small irregular pattern defects by com-paring Fourier spectra between the image and a template, showing the effectiveness of this method. While inspired by the idea of using templates, this approach differs from the proposed method by not utilizing Fourier spectrum comparison and incorporating the element of deep learning. J. Si et.al.[11] serves as a preliminary study for the proposed method, introducing the ability to detect defects by applying Fourier transform to the results of an autoencoder and removing specific components. Unlike [10], which used templates, this research demonstrates that improved performance can be achieved by solely focusing on component removal. Consequently, this paper introduces additional methods and achieves performance enhancement for various textures.\\
DRAEM[12] presents a defect detection study based on reconstructing anomalous data, which deviates from the conventional approach of training on normal data for reconstruction. This method simultaneously learns two networks for reconstruction and discrimination to preserve and detect defective regions. However, the goal of this paper is to achieve performance improvement using a simple approach by training only on normal data without generating anomalous data, thus it may achieve lower performance compared to the study mentioned. Y. Liang et.al.[13] mentions the limitations of reconstruction capabilities for other methods and introduces a method for defect detection from a frequency perspective, which aligns with the viewpoint and approach of this paper. Two novel methods, Frequency Decoupling and Channel Selection, are proposed to reconstruct from various frequency perspectives and combine them for more accurate defect detection. N-Pad[14] introduces a method for defect detection using relative positional information for each pix-el. The relative positional information is represented in eight directions, and through the use of a loss function, the paper demonstrates the utilization of this positional information. Anomaly Score is proposed using Mahalanobis and Euclidean distances, and various experiments on neighborhood sizes demonstrate the significance of the method.\\
J. Si et.al.[15] focuses on the application of reconstruction to thermal images of solar panels for defect detection. As the distribution of thermal images is sensitive to color and lacks pronounced edge features, this paper proposes a method using patches instead of reconstructing the entire image. The proposed method introduces a technique called "Difference Image Alignment Technique" by sorting pixel values, which enables easy detection of defects using only a few specific pixels. However, due to the significant differences in data characteristics between the focus of this paper (manufacturing) and thermal images, the application may not be straightforward. C. C. Tsai et.al.[16] introduces a method for defect detection by considering the similarity between patches in order to extract representative and important information from images. By utilizing different-sized patches, the method performs representation learning based on different scales and applies K-means clustering and cosine similarity for better defect detection. The advantage of randomly selecting multiple patches from the image and including local information through relative angles is demonstrated. However, while both object and texture were detected in this study, it showed lower performance specifically for texture detection, whereas this paper focuses solely on texture.\\
T. Liu et.al.[17] proposes a method for enhancing defect detection performance in grayscale images by applying post-processing techniques such as color space and image processing. To avoid incorrect classification of color information, the network is designed to reconstruct the original colors using grayscale images. By incorporating various augmentation techniques and morphology, the paper shows improved performance. Y. Shi et.al.[18] stands out from most other studies that focus on reconstructing images. Instead, this study utilizes a pretrained model to extract feature maps from various layers, combines them, and performs reconstruction to better restore features. By basing all the content on diverse feature maps, the method can better preserve defective regions in the results. H. Jinlei et.al.[17], a divide-and-assemble approach is proposed to overcome the limitations of AutoEncoder models in unsupervised anomaly detection. By applying this approach, the reconstruction capability of the model is modulated. The paper introduces a multi-scale block-wise memory module, adversarial learning, and meaningful latent representations to improve the performance of anomaly detection. The results demonstrate enhanced anomaly detection performance.

\section{Texture Defect Detection}
In this paper, we propose a defect detection method using deep learning networks and Fourier transformation. We define \textbf{TFR}, which stands for \textbf{T}exture Defect Detection with \textbf{F}ourier Transform using Normal \textbf{R}econstructed Template of Simple Autoencoder as the title of our paper. The proposed method follows the following steps:

\paragraph{Generation of Reconstructed Images through Denoising}: Initially, a deep learning network is employed to perform a de-noising process on the input images. This process generates reconstructed images where fine details have been removed. The network used in this step is a simple autoencoder with a straightforward structure, trained only on normal images. Its purpose is to generate images that closely resemble the input, primarily focusing on noise removal. The simple autoencoder is not involved in the task of defect detection.

\paragraph{Preservation Defect using Fourier Transformation}: Defect detection occurs during the testing phase. The trained model is utilized to create Normal Reconstructed Templates from a set of normal experimental data. One template is selected, and the same Fourier transformation process is applied to it. Defective regions in the Fourier domain differ from normal regions and are mainly preserved in the high-frequency components. Thus, some low-frequency components are removed, followed by inverse transformation.

\paragraph{Generation of Difference Images and Binary-Level Thresholding}: The inverse transformed result of the Normal Reconstructed Template is compared to the inverse transformed input image to obtain a difference image. This difference image represents the discrepancies between the defective and normal regions. By applying a threshold, the difference image is converted into a binary image. In the binary image, pixels corresponding to defects exhibit significant differences compared to using a normal image as input.

By following the aforementioned steps, the proposed method enables defect detection. It allows for the differentiation between normal and defective images, highlighting the regions where defects are present. \\

\begin{figure}[htb!]
    \centering
    \includegraphics[width=16cm]{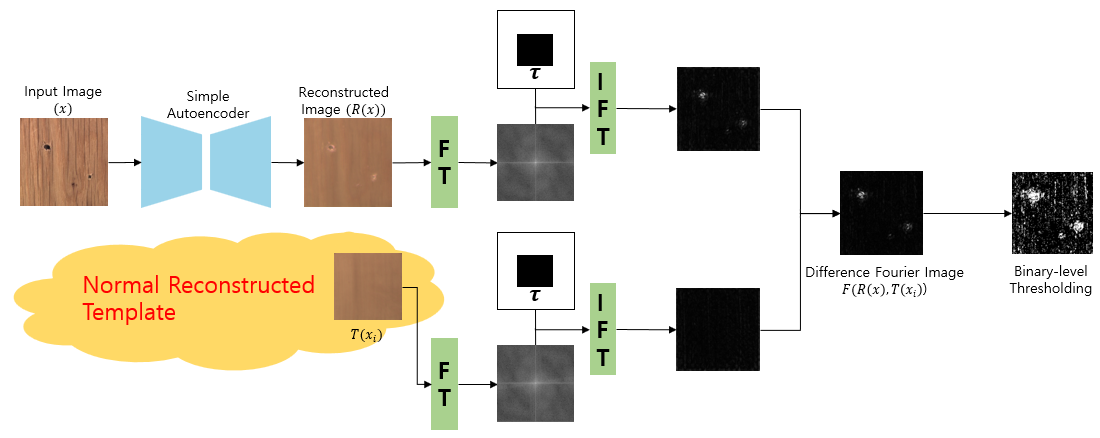}
    \caption{Overall architecture of proposed method}
\end{figure}

\subsection{Normal Texture Image Reconstruction}

This model processes input images of size 256x256, which are composed of RGB color channels. Therefore, the input shape is (256, 256, 3), and it consists of 5 layers in depth. This proposed network is showed in Fig. 2.\\
Autoencoder is composed of an encoder and a decoder. The input data is compressed into a low-dimensional latent vector through the encoder and then reconstructed back to the original size through the decoder. In other words, it generates an output of the same size as the input image.
The encoder part follows the structure of a Convolutional Neural Network (CNN). The first convolutional layer uses 64 filters, each applying a 3x3 kernel. It uses the ReLU activation function and applies 'same' padding to maintain the output size the same as the input. Then, additional Convolution layers are used to extract spatial features of the image and gradually reduce the size. The output of the encoder is passed to the decoder for the restoration process. The decoder uses up-sampling layer alternately to match the size of each layer mentioned in the encoder. It also incorporates a skip-connection structure, where the information from each layer of the encoder is brought and combined. As a result, it generates high-quality output results of the same size as the input image.\\
This model combines the basic L1 and L2 losses as its loss function. The L1 loss calculates the absolute error between the actual values and the predicted values, while the L2 loss calculates the squared error between them. By combining these two losses, the reconstruction loss (1) is computed and minimized during the model's training process. This combined form of a simple loss function provides an approach to capturing various aspects of the error in a balanced manner. By minimizing this combined loss, the model can effectively learn and optimize its parameters.\\

\begin{equation}
\mathcal L_{recon}(x,R(x)) = \lambda_{L1} \cdot L_1(x,R(x)) + \lambda_{L2} \cdot L_2(x,R(x))
 \label{eq1}
\end{equation}

In this paper, this simple autoencoder structure is used for defect detection tasks. The goal is to perform defect detection using this simple autoencoder structure. By effectively compressing and reconstructing the input data, autoencoders can detect and differentiate between normal and defective data.\\
Furthermore, this structure provides denoising effects. In the encoder part, it extracts features from the input image and goes through a process of removing noise. This helps reduce the noise in the input data. Here, noise refers to parts that can be misclassified as defects in the texture, such as patterns in the background of normal images. By removing noise, the reconstructed image should have a clean background. This is very useful when performing Fourier transformation for frequency band division. With reduced noise, the input data represents clearer and more accurate frequency bands, making it easier to distinguish defect areas from the results of Fourier transformation. Therefore, this simple autoencoder structure with denoising effects is highly suitable for Fourier transformation and can be effectively used for tasks such as defect detection and frequency band division.

\begin{figure}[htb!]
    \centering
    \includegraphics[width=16cm]{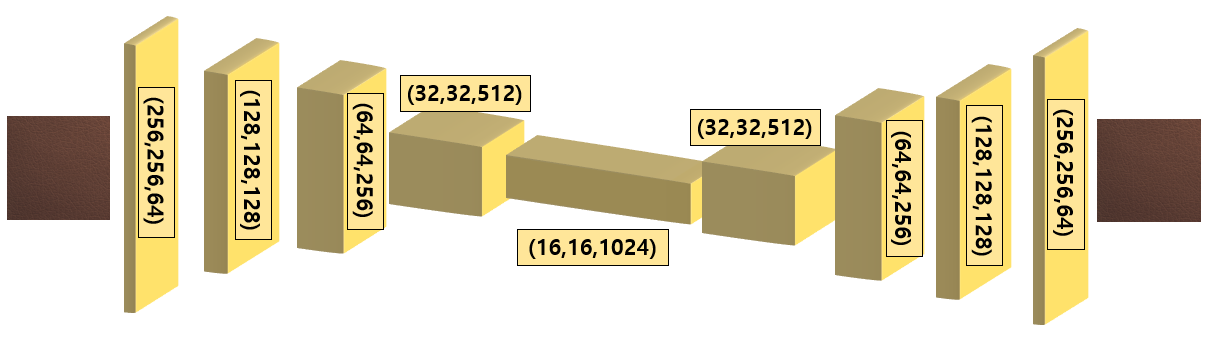}
    \caption{Model architecture for reconstruction}
\end{figure}

\subsection{Application of Fourier Transform}
The Fourier transform of a 2D image refers to the process of transforming the image from the spatial domain to the frequency domain, allowing us to obtain information related to the frequency components of the image. The Fourier transform is used in image processing and analysis by converting the original image to the frequency domain, performing necessary operations, and then restoring it back to the original domain through inverse transformation.\\
First, the given 2D square image (256, 256) is represented in the spatial domain using $(x,y)$ coordinates. To perform the Fourier transform on this image, we use Eq. (2): $F(u,v)$ is a complex number representing the transformed result in the frequency domain, and $(u,v)$ represents the coordinates in the frequency domain. Eq. (2) implies multiplying the complex exponential function in the frequency domain and the image value at each position in the spatial domain, and then summing them up for all positions. This allows us to obtain frequency information of the image in the frequency domain.\\

\begin{equation}
F(u,v) = \sum_{x=0}^{N-1} \sum_{y=0}^{N-1} f(x,y) \cdot e^{-j2\pi \left(\frac{ux+vy}{N}\right)}
 \label{eq2}
\end{equation}

The inverse Fourier transform is represented by Eq. (3). Importantly, the property of the Fourier transform is that when the original image is transformed and then inverse transformed, it is restored to the original image. This represents the relationship between the Fourier transform and the inverse Fourier transform.

\begin{equation}
f(x,y) = \frac{1}{N^2} \sum_{u=0}^{N-1} \sum_{v=0}^{N-1} F(u,v) \cdot e^{j2\pi \left(\frac{ux+vy}{N}\right)}
\label{eq3}
\end{equation}

In this paper, multiple reconstructed results for a "Normal Reconstructed Template" representing normal data are defined as $T(x_i)$. To determine the presence of defects, both the target image to be evaluated and the Normal Reconstructed Template are Fourier transformed to convert them into the frequency domain. In the frequency domain, the part with frequency 0 is placed at the center, and as the frequency increases, it is shifted towards the edges of the frequency area through a shift process.\\
Next, a "Fourier Mask" is defined to remove the low-frequency components, as shown in Fig. 3. For this purpose, a square mask with a side length of $\tau$ centered at the origin is created. This mask is used to perform pixel-wise operations with the Fourier transformed result, effectively removing the low-frequency components. This process retains only the high-frequency region where defects exist, while eliminating the background and unwanted components. Finally, an inverse transformation is applied to convert the result back from the frequency domain to the spatial domain. As a result, only the defective regions of the Texture image are preserved in the spatial domain.

\begin{figure}[htb!]
    \centering
    \includegraphics[width=16cm]{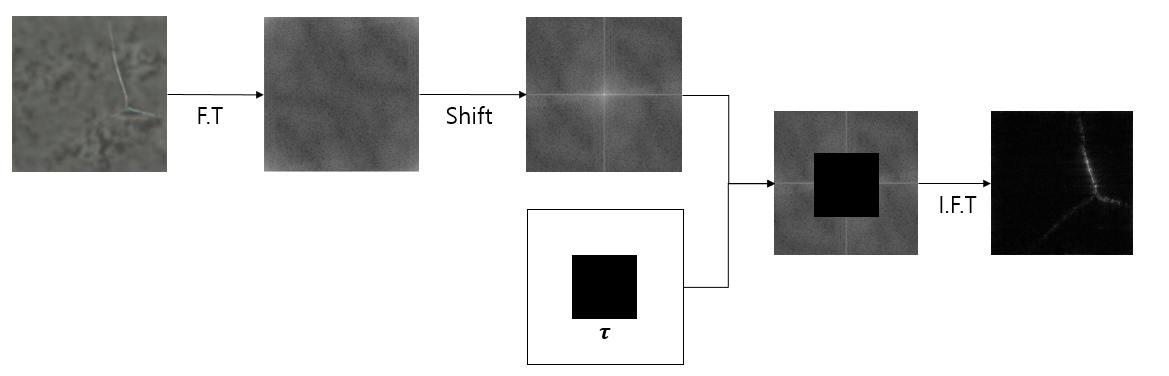}
    \caption{Fourier Mask and Pixel-wise}
\end{figure}

\subsection{Difference Images and Thresholding}
The two generated images ($F(R(x))$, $F(T(x_i))$) obtained through the given process represent the result of removing the low-frequency band and retaining only the high-frequency band. Therefore, the Normal Reconstructed Template retains only fine high-frequency details while removing the rest. If we perform the same process using a normal image as input, the difference with the Normal Reconstructed Template will be very small. However, when an image with defects is used as input, the defective parts are not completely eliminated through the process, and some noise from the background may remain. By subtracting the two generated images, the resulting difference image will have non-zero values in the defective regions and values close to zero in the remaining areas. This difference image can be used to create a final map for defect detection. Finally, by applying a threshold value $th$ to the generated map, we can generate final maps that allow us to determine the presence of defects. If the values exceed the threshold, they are set to one. Otherwise, they are set to zero. This binary map indicates the presence of defects where it is set to one and absence where it is set to zero. By calculating the total count of pixels in the generated map, we can derive scores for each image. Since the score range can vary significantly, we normalize the scores based on the scores of all the images and calculate an appropriate defect score.\\
Therefore, the resulting binary image obtained through this process will have one in the locations of defects and zero in the unaffected areas. This enables defect detection, and by calculating scores for the entire image and normalizing them, we can determine the normalized defect score. Fig. 4 represents the process of calculating the defect score.

\begin{figure}[htb!]
    \centering
    \includegraphics[width=16cm]{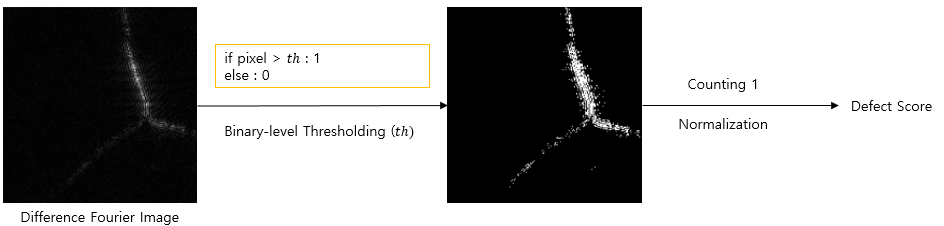}
    \caption{Process of calculating the defect score}
\end{figure}

\section{Experimental Results}
\subsection{Datasets}

In this paper, we focus on Texture Defect Detection using the MV-Tec AD[20]. This datasets consist of 5 textures and 10 objects, but we only evaluate the performance using 5 textures. However, the available data is insufficient for both training and testing. Therefore, data augmentation is performed in this paper. Tab. 1 represents the final composition of the datasets in terms of the number of samples for each category. Since we only use normal data for training, the majority of the existing data is used as training data. To balance the number of samples between normal and defect data, augmentation techniques are applied to the normal data. This process increases the diversity of the training data and ensures an adequate number of defect samples, ultimately improving the performance of the model.

\begin{table}[htbp]
 \caption{Detailed datasets with data augmentation}
  \centering
  \begin{tabular}{lrrrr}
    \toprule
    Category  &Train(Normal) &Test(Normal) &Test(Defect)\\
     \midrule
    Carpet &280 &84 &89\\
    Grid &264 &63 &57\\
    Leather &245 &96 &92\\
    Tile &230 &99 &84\\
    Wood &247 &57 &60\\
   
    \bottomrule
  \end{tabular}
  \label{tab:table}
\end{table}

\subsection{Training Details}

For Texture Defect Detection in this study, the following training approach is utilized. The input normal image data is normalized within the range of 0 and 1, and data augmentation is performed using ImageDataGenerator to generate diverse forms of training data. Data augmentation techniques such as shearing (20\%), zooming (20\%), and vertical and horizontal flipping are applied. During the training process, the Adam optimizer is employed with an initial learning rate set to 1e-4. The training is conducted for 500 epochs on the entire datasets, with a batch size of 16. In the loss function, the hyperparameter $\lambda_{L2}$ is set to 100 for L2 loss, and $\lambda_{L1}$ is set to 1 for L1 loss, resulting in a combination of simple loss functions.

\subsection{Performance Evaluation and Ablation Study}

The proposed network in this study is trained only on normal data, resulting in the generation of images with different distributions when dealing with defect images. Fig. 5 illustrates the inferred reconstructed images from examples of defective data for each category. All the data in the figure contains defects. The first row shows the original images with defects, while the second row represents the reconstructed images. The original images exhibit patterns even in the background, indicating prominent features. However, the defective regions possess even more pronounced features. Therefore, by removing the noise in the background, the defective regions can be highlighted more effectively. The overall reconstructed results appear blurred, with a significant reduction in noise in the background except for the grid. Although the defective regions also become blurry, the removal of background patterns makes it easier to extract the defects. The grid exhibits a consistent pattern and closely resembles the original, except for the areas with defects where differences can be observed.

\begin{figure}[htb!]
    \centering
    \includegraphics[width=16cm]{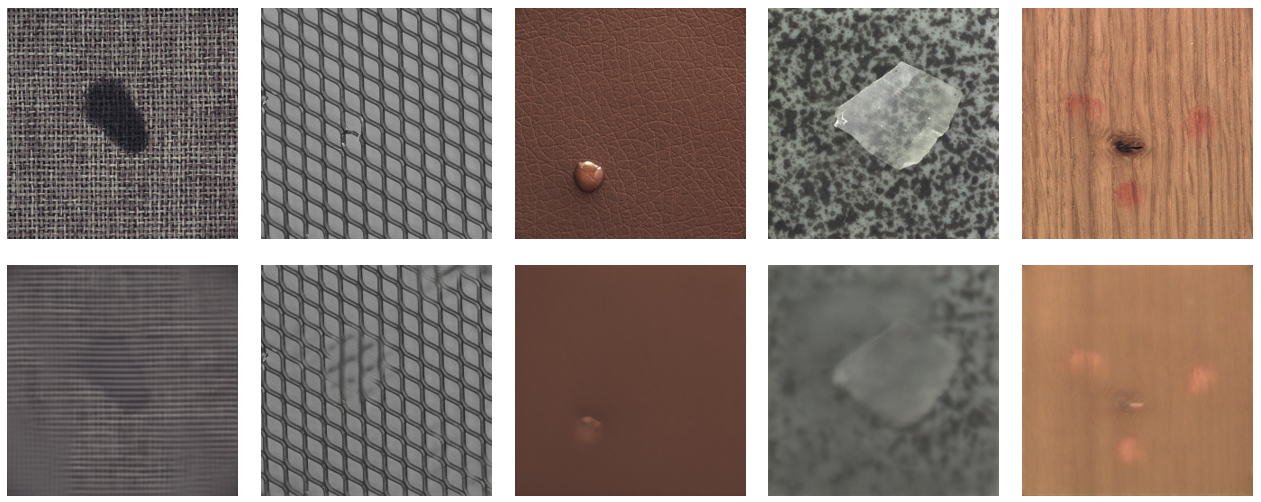}
    \caption{Reconstructed image with defect}
\end{figure}

The reason for using Normal Reconstructed Templates is to overcome the limitations of the network and improve performance by assessing the differences between the closest reconstructed image and the original image, even for normal data. In reality, even normal data cannot be perfectly reconstructed to match the original. Hence, the approach involves extracting the normal regions by utilizing the differences between the reconstructed image and the original image as closely as possible. Normal Reconstructed Template represents the restoration results of normal data and possess various forms and patterns. These templates are used to generate difference Fourier images, which are then employed to detect defective regions. As the network is trained solely on normal data, processing data with defects will result in reconstructed images with slightly different distributions. Thus, the approach involves calculating the differences between the normal reconstructed templates and the defective regions to extract the defects.\\
This approach allows for the precise detection of defects by distinguishing between normal and defective regions. By removing the normal regions based on the differences with the reconstructed templates, the remaining areas are composed of the defective regions, making the defects more distinct and enabling more accurate detection. Therefore, in this study, the proposal of utilizing normal reconstructed templates aims to overcome the limitations of the network and enhance performance. By removing the normal regions and emphasizing the defective regions, more accurate defect detection can be achieved. The reconstructed images from normal data exhibit a variety of normal restoration templates. Therefore, it is crucial to select the most suitable template for evaluating the reconstructed results on test normal images. Fig. 6 presents the selection of appropriate templates for each category based on experimental results. These templates can be utilized to generate difference Fourier images and improve performance consistently across all data. Additionally, when generating difference Fourier images, the 10-pixel edge is excluded from evaluation. This is because the edge exhibits a different distribution compared to the original due to padding, increasing the likelihood of misclassification.\\

\begin{figure}[htb!]
    \centering
    \includegraphics[width=16cm]{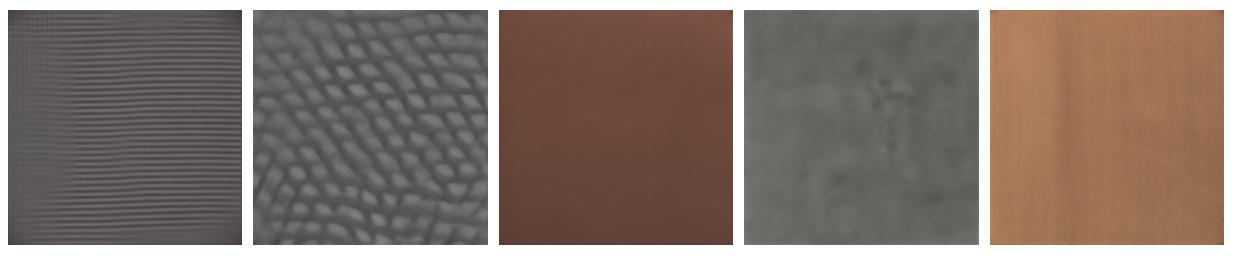}
    \caption{Normal Reconstructed Template}
\end{figure}

As mentioned in Chapter 3, we need to find the most suitable template by combining the length of the Fourier mask, denoted as $\tau$, and the binary-level thresholding represented by $th$. To evaluate the performance, we use the Area Under the Curve (AUC) as the evaluation metric. AUC is a common metric used to assess the performance of classification models, ranging from 0 to 1, where a value closer to 1 indicates better performance. Since each category has different characteristics, they have different parameter values, and we explore various combinations of these parameters. Therefore, we use AUC to find the optimal parameter combination for each category and select the combination with the highest AUC value. Based on this, we inferred the AUC values for various parameter combinations and presented the results in Table 2. The parameter combinations that showed the highest AUC for each category are as follows ($\tau$, $th$): Carpet: (40, 13), Grid: (43, 16), Leather: (3, 4), Tile: (40, 2), Wood: (2, 9).\\
The overall average AUC is 93.1\%. This indicates that our proposed simple method achieves performance similar to state-of-the-art approaches. These results demonstrate that our method effectively detects defects despite its simplicity. However, the Carpet category recorded a relatively lower AUC compared to other categories. This can be attributed to the relatively smaller difference between the defect regions and the background in the Carpet category compared to other categories. When the difference between defect regions and the background is small, removing noise from the background may also result in the removal of defect regions, making accurate detection more challenging. Therefore, the Carpet category may require different parameter values and additional adjustments.\\
In conclusion, we have shown that a simple method can achieve high performance. Additionally, since optimal parameter combinations may vary for each category, it is important to adjust parameter values accordingly. Thus, our method offers both flexibility and simplicity, making it applicable to various defect detection problems.\\
Fig. 7 shows a partial process of generating the final decision map for each category. The first column represents the reconstructed image. The second and third columns show frequency domain images of the input image and normal reconstructed template, respectively. The fourth column illustrates the result of binary-level thresholding applied to the difference Fourier image. The white areas indicate defective regions, and it can be observed that the actual defective areas are well preserved.

\begin{table}[htbp]
\caption{Performance analysis (row: Fourier mask length $\tau$, column: threshold value $th$) }
\centering
\begin{tabular}{c|cccccc}
\hline
Carpet & 9 & 10 & 11 & 12 & 13 & 14 \\
\hline
37 & 0.831 & 0.790 & 0.753 & 0.720 & 0.687 & 0.679 \\
38 & 0.771 & 0.751 & 0.733 & 0.719 & 0.705 & 0.697 \\
39 & 0.798 & 0.779 & 0.767 & 0.761 & 0.764 & 0.762 \\
40 & 0.852 & 0.848 & 0.844 & 0.852 & \textbf{0.854} & 0.829 \\
41 & 0.836 & 0.825 & 0.796 & 0.796 & 0.728 & 0.663 \\
42 & 0.796 & 0.768 & 0.752 & 0.715 & 0.645 & 0.566 \\
\hline\hline

Grid & 15 & 16 & 17 & 18 & 19 & 20 \\
\hline
42 & 0.917 & 0.906 & 0.929 & 0.926 & 0.923 & 0.911 \\
43 & 0.913 & \textbf{0.934} & 0.927 & 0.925 & 0.920 & 0.912 \\
44 & 0.920 & 0.923 & 0.926 & 0.926 & 0.918 & 0.884 \\
45 & 0.924 & 0.927 & 0.918 & 0.914 & 0.893 & 0.875 \\
46 & 0.916 & 0.923 & 0.918 & 0.915 & 0.901 & 0.884 \\
47 & 0.913 & 0.912 & 0.901 & 0.899 & 0.896 & 0.873 \\
\hline\hline

Leather & 1 & 2 & 3 & 4 & 5 & 6 \\
\hline
1 & 0.781 & 0.728 & 0.736 & 0.843 & 0.926 & 0.961 \\
2 & 0.746 & 0.792 & 0.882 & 0.935 & 0.926 & 0.868 \\
3 & 0.752 & 0.865 & 0.946 & \textbf{0.974} & 0.916 & 0.871 \\
4 & 0.633 & 0.818 & 0.908 & 0.931 & 0.900 & 0.855 \\
5 & 0.562 & 0.794 & 0.916 & 0.906 & 0.897 & 0.853 \\
6 & 0.513 & 0.768 & 0.891 & 0.916 & 0.890 & 0.835 \\
\hline\hline

Tile & 1 & 2 & 3 & 4 & 5 & 6 \\
\hline
37 & 0.729 & 0.907 & 0.752 & 0.95 & 0.643 & 0.607 \\
38 & 0.729 & 0.893 & 0.744 & 0.687 & 0.643 & 0.607 \\
39 & 0.746 & 0.920 & 0.749 & 0.687 & 0.637 & 0.607 \\
40 & 0.747 & \textbf{0.921} & 0.949 & 0.681 & 0.625 & 0.601 \\
41 & 0.761 & 0.919 & 0.739 & 0.675 & 0.619 & 0.601 \\
42 & 0.762 & 0.892 & 0.737 & 0.675 & 0.613 & 0.601 \\
\hline\hline

Wood & 7 & 8 & 9 & 10 & 11 & 12 \\
\hline
1 & 0.834 & 0.835 & 0.860 & 0.898 & 0.913 & 0.919 \\
2 & 0.899 & 0.925 & \textbf{0.971} & 0.961 & 0.934 & 0.919 \\
3 & 0.916 & 0.946 & 0.938 & 0.938 & 0.928 & 0.901 \\
4 & 0.923 & 0.921 & 0.926 & 0.918 & 0.917 & 0.911 \\
5 & 0.929 & 0.911 & 0.909 & 0.905 & 0.908 & 0.900 \\
6 & 0.895 & 0.900 & 0.897 & 0.906 & 0.897 & 0.884 \\
\hline

\end{tabular}
\label{table:carpet-auc}
\end{table}

\begin{figure}[htb!]
    \centering
    \includegraphics[width=16cm]{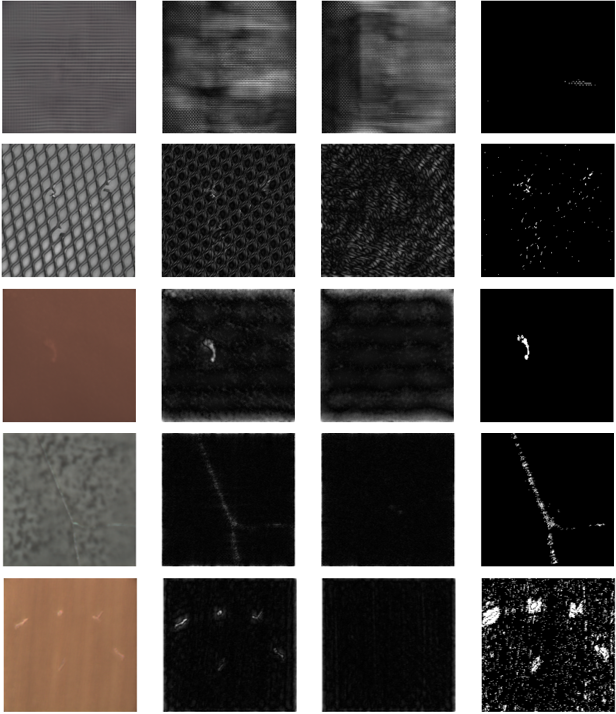}
    \caption{Process for locating defect detection (1st column: reconstruction, 2nd-3rd columns: Fourier trans-form results, 4th column: binary-level thresholding)}
\end{figure}

Tab. 3 presents the results of evaluating various anomaly detection algorithms on different categories. The performance of each algorithm is measured by the AUC (Area Under the Curve) value, where a value of 1 indicates the presence of defects and 0 indicates their absence.\\
Analyzing the results, we can observe that the algorithm DAAD[19] achieved the highest AUC value of 0.866 for the Carpet category. This indicates that DAAD[19] demonstrates effective performance in anomaly detection for Carpet data. Additionally, for the Grid and Wood category, DAAD[19] also shows a high AUC value of 0.957 and 0.982, indicating its superior performance in anomaly detection. For the Leather category, the proposed method achieves the highest AUC value of 0.974. This suggests that proposed method have strong anomaly detection performance for Leather data, respectively. For the Tile category, proposed methods obtain the highest AUC value of 0.921. This indicates that the proposed method demonstrates excellent anomaly detection capability for Tile data.\\
Overall, AnoGAN[1] shows relatively lower performance across all categories, while GANomaly[3] and Skip-GANomaly[4] exhibit highly irregular results compared to other categories. DAAD[19] demonstrates high performance for specific categories, but the proposed method shows excellent performance in all categories except Carpet. Notably, the proposed method stands out as it does not heavily rely on deep learning networks compared to previous studies, and it exhibits the smallest category-wise variation with the highest average AUC of 93.1\%. The significant difference observed in the Leather category can be attributed to the simplicity of the background, making noise removal easier.

\begin{table}[htbp]
\caption{: Performance analysis (row: Fourier mask length $\tau$, column: threshold value $th$) }
\centering
\begin{tabular}{l|ccccccc|}
\hline
Methods & AnoGAN[1] & memAE[5] & OCGAN[6] & GANomaly[3] & Skip-GANomaly[4] & DAAD[19] & Ours \\
\hline
Carpet & 0.337 & 0.386 & 0.348 & 0.699 & 0.795 & \textbf{0.866} & \underline{0.854} \\
Grid & 0.871 & 0.805 & 0.855 & 0.708 & 0.657 & \textbf{0.957} & \underline{0.934} \\
Leather & 0.451 & 0.423 & 0.624 & 0.842 & \underline{0.908} & 0.862 & \textbf{0.974} \\
Tile & 0.401 & 0.718 & 0.806 & 0.794 & 0.850 & 0.882 & \textbf{0.921} \\
Wood & 0.567 & 0.954 & 0.959 & 0.834 & 0.919 & \textbf{0.982} & \underline{0.971} \\
\hline
Average & 0.525 & 0.657 & 0.718 & 0.775 & 0.826 & \underline{0.910} & \textbf{0.931} \\
\hline
\end{tabular}
\label{table:method-comparison}
\end{table}

The proposed method in this paper has two main aspects: using a simple autoencoder for re-construction and combining it with Fourier transform to separate and remove defective features. Therefore, the performance is evaluated based on the usage of reconstruction and Fourier trans-form. Without reconstruction, performing only Fourier transform makes it extremely challenging to separate defective features in the frequency domain, resulting in a slightly higher AUC of 53.1\% overall compared to random results. When only reconstruction is used along with Fourier transform, it yields very low performance with an average AUC of 63.5\% across all categories except for Tile. This indicates that using only reconstruction to preserve defective regions is difficult for other categories, except for Tile, where the reconstructed results alone can preserve defective regions. On the other hand, the proposed method that combines both reconstruction and Fourier transform achieves the highest performance with an AUC of 93.1\%. Although there is a 2.6\% AUC decrease for Tile, there is an overall improvement of up to 46.6\%, making the proposed method valid.

\begin{table}[htbp]
\caption{: Performance analysis (row: Fourier mask length $\tau$, column: threshold value $th$) }
\centering
\begin{tabular}{l|ccc}
\hline
 Reconstruction &X &O &O\\
 Fourier Transform &O &X &O\\
\hline
Carpet & 0.506 & 0.697 & \textbf{0.854} \\
Grid & 0.634 & 0.508 & \textbf{0.934} \\
Leather & 0.505 & 0.517 & \textbf{0.974} \\
Tile & 0.505 & \textbf{0.947} & 0.921 \\
Wood & 0.503 & 0.505 &\textbf{0.971} \\
\hline
Average & 0.531 & 0.635 & \textbf{0.931} \\
\hline
\end{tabular}
\label{table:method-comparison}
\end{table}

\section{Conclusions}
This study proposes a method in the field of image analysis that combines a simple autoencoder and Fourier transform for texture defect detection. The experimental results showed that the pro-posed method achieved overall high performance, although there were cases where it did not show effectiveness for some data. Therefore, for future research, it is necessary to enhance the performance of the proposed method by utilizing deep learning networks to improve noise removal. By developing and applying more sophisticated denoising techniques through deep learning networks, the defect detection performance can be further improved. Additionally, future research directions include conducting comparative studies with other anomaly detection algorithms to validate the superiority of the proposed method and performing experiments on a wider range of datasets to evaluate its generalization performance.


\documentclass{article}


\usepackage{arxiv}

\usepackage[utf8]{inputenc} % allow utf-8 input
\usepackage[T1]{fontenc}    % use 8-bit T1 fonts
\usepackage{hyperref}       % hyperlinks
\usepackage{url}            % simple URL typesetting
\usepackage{booktabs}       % professional-quality tables
\usepackage{amsfonts}       % blackboard math symbols
\usepackage{nicefrac}       % compact symbols for 1/2, etc.
\usepackage{microtype}      % microtypography
\usepackage{lipsum}
\usepackage{graphicx}
\graphicspath{ {./images/} }


\title{TFR: Texture Defect Detection with Fourier Transform using Normal Reconstructed Template of Simple Autoencoder
}


\author{
 Jongwook Si \\
  Dept. Computer AI Convergence Engineering\\
  Kumoh National Institute of Technology\\
  Gumi, KOREA 39177 \\
  \texttt{jwsi425@kumoh.ac.kr} \\
  %% examples of more authors
   \And
 Sungyoung Kim \\
  Dept. Computer Engineering\\
  Kumoh National Institute of Technology\\
  Gumi, KOREA 39177 \\
  \texttt{sykim@kumoh.ac.kr} \\
  \And
}


\begin{document}
\maketitle
\begin{abstract}
Texture is an essential information in image representation, capturing patterns and structures. As a result, texture plays a crucial role in the manufacturing industry and is extensively studied in the fields of computer vision and pattern recognition. However, real-world textures are susceptible to defects, which can degrade image quality and cause various issues. Therefore, there is a need for accurate and effective methods to detect texture defects. In this study, a simple autoencoder and Fourier transform are employed for texture defect detection. The proposed method combines Fourier transform analysis with the reconstruct-ed template obtained from the simple autoencoder. Fourier transform is a powerful tool for analyzing the frequency domain of images and signals. Moreover, since texture defects often exhibit characteristic changes in specific frequency ranges, analyzing the frequency domain enables effective defect detection. The proposed method demonstrates effectiveness and accuracy in detecting texture defects. Experimental results are presented to evaluate its performance and compare it with existing approaches.
\end{abstract}


% keywords can be removed
\keywords{ Texture\and Defect Detection\and Anomaly Detection\and Fourier Transform\and Reconstruction}


\section{Introduction}
Texture defect detection is a crucial topic in image processing and computer vision, as texture provides important information about patterns and structures in images. Texture defects are particularly significant in the manufacturing industry, where they are essential for evaluating product quality and detecting defects in manufacturing processes.\\
In the manufacturing industry, it is crucial to ensure that the produced products meet the required standards. Texture defects can significantly impact product quality. Texture defect detection allows for the identification and resolution of anomalies or irregularities occurring in texture pat-terns or structures, enabling the evaluation of product quality. The paper focuses on proposing a new approach that combines Fourier transform and autoencoder-based methods to effectively detect and address texture defects.\\
Texture defects can arise during the manufacturing process, and detecting and identifying such defects is essential to maintain accuracy. By applying texture defect detection techniques, manufacturing companies can swiftly detect issues and take prompt actions. Furthermore, texture defect detection offers advantages in terms of cost reduction. Texture defects can lead to additional costs for manufacturing companies, such as customer dissatisfaction. By implementing robust texture defect detection methods, companies can minimize the occurrence of defective products, reduce waste, and optimize production efficiency.\\
Manual inspection of texture defects is time-consuming and subjective, posing limitations. The automation of the texture defect detection process using computer vision technology enables faster and more objective evaluations. The paper aims to contribute to the improvement of product quality, optimization of manufacturing processes, cost reduction, and automation by advocating the importance of texture defect detection and proposing a novel approach that integrates Fourier trans-form and autoencoder-based methods.\\
Various techniques have been used in texture defect detection, including the combination of deep learning and image processing methods. In this study, Fourier transform and a simple autoencoder are utilized for texture defect detection. Fourier transform is a useful tool for analyzing the frequency domain of images or signals, enabling the characterization of texture frequency proper-ties. Since texture defects may exhibit characteristic changes in certain frequency ranges, analyzing the frequency domain using Fourier transform allows for effective defect detection. Additionally, the autoencoder is employed to learn the normal characteristics of textures and reconstruct them. An autoencoder is a model that encodes input data into a latent space and reconstructs it, thereby extracting features from the input data. In this research, the autoencoder is used to reconstruct nor-mal texture data and generate a template of normal reconstruction. This template is utilized for de-fect detection by analyzing the differences between the reconstructed texture and the normal state. This paper provides a detailed explanation of the reconstruction process of the autoencoder and the method for generating the normal reconstruction template for texture defect detection. The aim is to contribute a fresh perspective and advancements to the field of texture defect detection. \\

The contributions of this paper are as follows:\\

•	Performance Improvement: The proposed simple autoencoder architecture achieves a high level of performance. Despite its simplicity, the autoencoder demonstrates effectiveness in texture defect detection, proving that efficient defect detection can be achieved without complex deep learning models.\\
•	Integration of Techniques: This study combines deep learning and image processing techniques for texture defect detection. Deep learning is used for denoising and reconstruction, while image processing methods are employed for extracting texture features and defect detection. This combination offers advantages such as not requiring extensive data training. Thus, the research presents a concise and effective approach for texture defect detection by integrating deep learning and image processing techniques.\\
•	Experimentation and Analysis: Detailed experiments and analyses are conducted on the necessary parameters. Various parameters are adjusted and compared to optimize the performance of texture defect detection. This provides insights into the parameters that have the most significant impact on performance and offers practical guidelines for real-world applications.\\

Through these contributions, this paper demonstrates the potential for performance enhance-ment in texture defect detection and the benefits of combining deep learning and image processing techniques.


\section{Related Works}
\label{sec:headings}
In anomaly detection research, methods based on reconstruction are widely studied. These methods typically involve training on normal data to generate reconstructed data and utilize the difference between the input anomalous data and the original image for detection. Models such as AutoEncoder and GAN are commonly employed for reconstruction and leveraging reconstruction errors for anomaly detection.


\subsection{Anomaly Detection with Reconstruction}

AnoGAN[1] proposes a basic approach to anomaly detection by combining unsupervised learning with GAN. It learns the distribution of normal data by inputting only normal data and calculates anomaly scores to compare and detect anomalies. f-AnoGAN[2], an extension of AnoGAN[1], improves performance by introducing fast mapping techniques for new data and incorporating an Encoder into GAN for more refined reconstruction. The generated data by f-AnoGAN[2] exhibits high generation performance that even experts find it difficult to distinguish from real data. GANomaly[3] is a method that learns both generation and latent space by using only normal data. Anomaly scores are computed based on the differences in latent vectors. Skip-GANomaly[4] ex-tends GANomaly[3] with a U-Net-based network architecture and introduces adversarial training that includes a loss function for Discriminator's feature maps, leading to improved reconstruction performance. MemAE [5] improves the limitations of using AutoEncoder for anomaly detection through the incorporation of a Memory Module, which makes reconstruction more challenging for abnormal samples. While AutoEncoder generalizes well, it can also reconstruct abnormal regions, which is a drawback. This paper focuses on post-processing methods to address this issue. OCGAN [6] is a model for one-class anomaly detection, where it learns latent representations of in-class examples and restricts the latent space to the given class. By utilizing a denoising AutoEncoder net-work and a discriminator, it generates in-class samples and explores anomalies outside the class boundaries. This approach achieves high-performance results.\\
These studies[1-6] mainly focus on reconstruction-based methods for anomaly detection, where training on normal data is used to assess and detect anomalies. Anomaly detection encompasses various subfields[7-9] that can be applied in real-life scenarios, including disease detection, accident detection, and fall detection, among others.\\
Y. Zhao et.al.[7]’s the objective is to effectively detect diseases in plants. However, the paper mentions the issue of data imbalance between diseased and healthy samples and proposes a solution called DoublaGAN, which applies Super-Resolution to augment the diseased data. This re-search achieves the goal of detecting various plant diseases while enhancing the resolution by a factor of four and demonstrates high performance. J. Si et.al.[8] introduces the structure of a Generative Adversarial Serial Autoencoder, which consists of Autoencoders connected in series. The pa-per presents a method for detecting diseases in chili peppers by using the Grabcut technique to segment the pepper regions and applying a reconstruction process. It addresses the limitations of reconstruction and improves performance by calculating all scores based on reconstruction. J. Si et.al.[9] is focused on detecting traffic accidents in black box videos. The proposed method generates the next frame of the video using information from the previous frames and compares it with the actual frame to detect accidents. However, the paper mentions the need to address the issue of misclassification in the background when dealing with moving videos.


\subsection{Defect Detection}

D. M. Tsai et.al.[10], a study proposing the use of Fourier transform to detect defects in PCBs is presented. The research demonstrates the ability to detect small irregular pattern defects by com-paring Fourier spectra between the image and a template, showing the effectiveness of this method. While inspired by the idea of using templates, this approach differs from the proposed method by not utilizing Fourier spectrum comparison and incorporating the element of deep learning. J. Si et.al.[11] serves as a preliminary study for the proposed method, introducing the ability to detect defects by applying Fourier transform to the results of an autoencoder and removing specific components. Unlike [10], which used templates, this research demonstrates that improved performance can be achieved by solely focusing on component removal. Consequently, this paper introduces additional methods and achieves performance enhancement for various textures.\\
DRAEM[12] presents a defect detection study based on reconstructing anomalous data, which deviates from the conventional approach of training on normal data for reconstruction. This method simultaneously learns two networks for reconstruction and discrimination to preserve and detect defective regions. However, the goal of this paper is to achieve performance improvement using a simple approach by training only on normal data without generating anomalous data, thus it may achieve lower performance compared to the study mentioned. Y. Liang et.al.[13] mentions the limitations of reconstruction capabilities for other methods and introduces a method for defect detection from a frequency perspective, which aligns with the viewpoint and approach of this paper. Two novel methods, Frequency Decoupling and Channel Selection, are proposed to reconstruct from various frequency perspectives and combine them for more accurate defect detection. N-Pad[14] introduces a method for defect detection using relative positional information for each pix-el. The relative positional information is represented in eight directions, and through the use of a loss function, the paper demonstrates the utilization of this positional information. Anomaly Score is proposed using Mahalanobis and Euclidean distances, and various experiments on neighborhood sizes demonstrate the significance of the method.\\
J. Si et.al.[15] focuses on the application of reconstruction to thermal images of solar panels for defect detection. As the distribution of thermal images is sensitive to color and lacks pronounced edge features, this paper proposes a method using patches instead of reconstructing the entire image. The proposed method introduces a technique called "Difference Image Alignment Technique" by sorting pixel values, which enables easy detection of defects using only a few specific pixels. However, due to the significant differences in data characteristics between the focus of this paper (manufacturing) and thermal images, the application may not be straightforward. C. C. Tsai et.al.[16] introduces a method for defect detection by considering the similarity between patches in order to extract representative and important information from images. By utilizing different-sized patches, the method performs representation learning based on different scales and applies K-means clustering and cosine similarity for better defect detection. The advantage of randomly selecting multiple patches from the image and including local information through relative angles is demonstrated. However, while both object and texture were detected in this study, it showed lower performance specifically for texture detection, whereas this paper focuses solely on texture.\\
T. Liu et.al.[17] proposes a method for enhancing defect detection performance in grayscale images by applying post-processing techniques such as color space and image processing. To avoid incorrect classification of color information, the network is designed to reconstruct the original colors using grayscale images. By incorporating various augmentation techniques and morphology, the paper shows improved performance. Y. Shi et.al.[18] stands out from most other studies that focus on reconstructing images. Instead, this study utilizes a pretrained model to extract feature maps from various layers, combines them, and performs reconstruction to better restore features. By basing all the content on diverse feature maps, the method can better preserve defective regions in the results. H. Jinlei et.al.[17], a divide-and-assemble approach is proposed to overcome the limitations of AutoEncoder models in unsupervised anomaly detection. By applying this approach, the reconstruction capability of the model is modulated. The paper introduces a multi-scale block-wise memory module, adversarial learning, and meaningful latent representations to improve the performance of anomaly detection. The results demonstrate enhanced anomaly detection performance.


\section{Texture Defect Detection}
In this paper, we propose a defect detection method using deep learning networks and Fourier transformation. We define \textbf{TFR}, which stands for \textbf{T}exture Defect Detection with \textbf{F}ourier Transform using Normal \textbf{R}econstructed Template of Simple Autoencoder as the title of our paper. The proposed method follows the following steps:

\paragraph{Generation of Reconstructed Images through Denoising}: Initially, a deep learning network is employed to perform a de-noising process on the input images. This process generates reconstructed images where fine details have been removed. The network used in this step is a simple autoencoder with a straightforward structure, trained only on normal images. Its purpose is to generate images that closely resemble the input, primarily focusing on noise removal. The simple autoencoder is not involved in the task of defect detection.

\paragraph{Preservation Defect using Fourier Transformation}: Defect detection occurs during the testing phase. The trained model is utilized to create Normal Reconstructed Templates from a set of normal experimental data. One template is selected, and the same Fourier transformation process is applied to it. Defective regions in the Fourier domain differ from normal regions and are mainly preserved in the high-frequency components. Thus, some low-frequency components are removed, followed by inverse transformation.

\paragraph{Generation of Difference Images and Binary-Level Thresholding}: The inverse transformed result of the Normal Reconstructed Template is compared to the inverse transformed input image to obtain a difference image. This difference image represents the discrepancies between the defective and normal regions. By applying a threshold, the difference image is converted into a binary image. In the binary image, pixels corresponding to defects exhibit significant differences compared to using a normal image as input.

By following the aforementioned steps, the proposed method enables defect detection. It allows for the differentiation between normal and defective images, highlighting the regions where defects are present. \\

\begin{figure}[htb!]
    \centering
    \includegraphics[width=16cm]{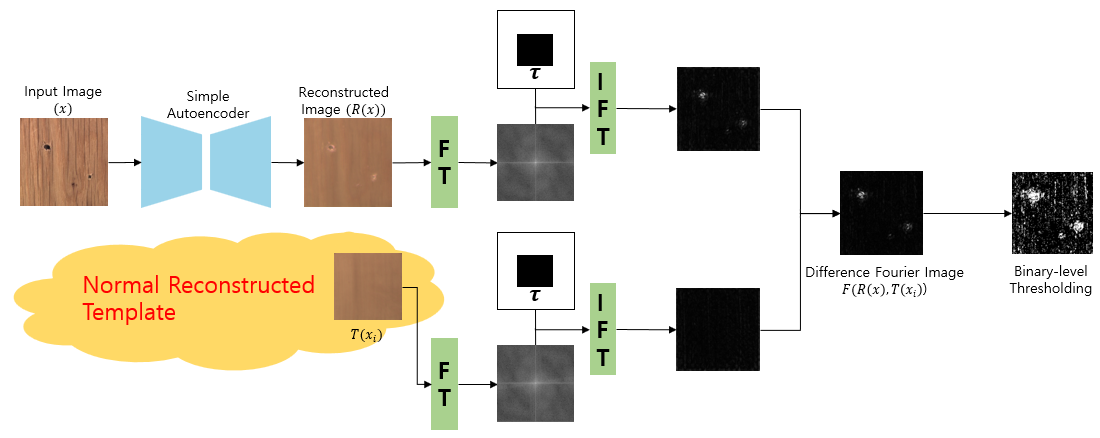}
    \caption{Overall architecture of proposed method}
\end{figure}


\subsection{Normal Texture Image Reconstruction}

This model processes input images of size 256x256, which are composed of RGB color channels. Therefore, the input shape is (256, 256, 3), and it consists of 5 layers in depth. This proposed network is showed in Fig. 2.\\
Autoencoder is composed of an encoder and a decoder. The input data is compressed into a low-dimensional latent vector through the encoder and then reconstructed back to the original size through the decoder. In other words, it generates an output of the same size as the input image.
The encoder part follows the structure of a Convolutional Neural Network (CNN). The first convolutional layer uses 64 filters, each applying a 3x3 kernel. It uses the ReLU activation function and applies 'same' padding to maintain the output size the same as the input. Then, additional Convolution layers are used to extract spatial features of the image and gradually reduce the size. The output of the encoder is passed to the decoder for the restoration process. The decoder uses up-sampling layer alternately to match the size of each layer mentioned in the encoder. It also incorporates a skip-connection structure, where the information from each layer of the encoder is brought and combined. As a result, it generates high-quality output results of the same size as the input image.\\
This model combines the basic L1 and L2 losses as its loss function. The L1 loss calculates the absolute error between the actual values and the predicted values, while the L2 loss calculates the squared error between them. By combining these two losses, the reconstruction loss (1) is computed and minimized during the model's training process. This combined form of a simple loss function provides an approach to capturing various aspects of the error in a balanced manner. By minimizing this combined loss, the model can effectively learn and optimize its parameters.\\

\begin{equation}
\mathcal L_{recon}(x,R(x)) = \lambda_{L1} \cdot L_1(x,R(x)) + \lambda_{L2} \cdot L_2(x,R(x))
 \label{eq1}
\end{equation}

In this paper, this simple autoencoder structure is used for defect detection tasks. The goal is to perform defect detection using this simple autoencoder structure. By effectively compressing and reconstructing the input data, autoencoders can detect and differentiate between normal and defective data.\\
Furthermore, this structure provides denoising effects. In the encoder part, it extracts features from the input image and goes through a process of removing noise. This helps reduce the noise in the input data. Here, noise refers to parts that can be misclassified as defects in the texture, such as patterns in the background of normal images. By removing noise, the reconstructed image should have a clean background. This is very useful when performing Fourier transformation for frequency band division. With reduced noise, the input data represents clearer and more accurate frequency bands, making it easier to distinguish defect areas from the results of Fourier transformation. Therefore, this simple autoencoder structure with denoising effects is highly suitable for Fourier transformation and can be effectively used for tasks such as defect detection and frequency band division.

\begin{figure}[htb!]
    \centering
    \includegraphics[width=16cm]{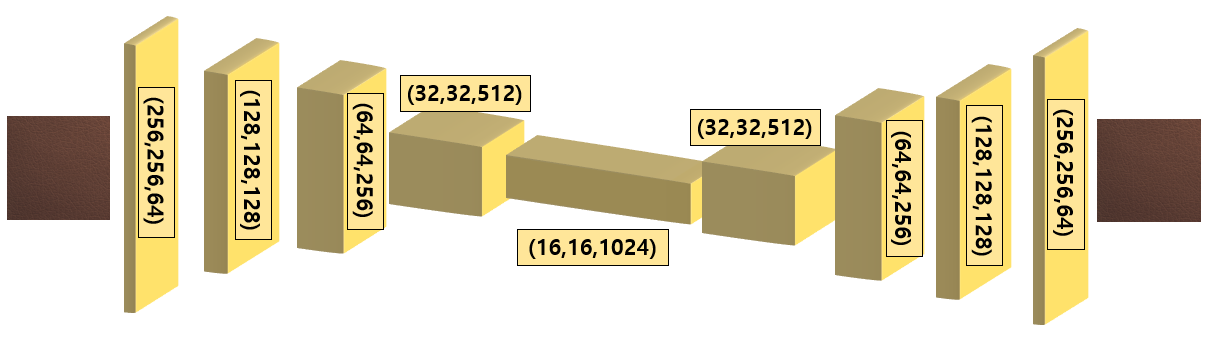}
    \caption{Model architecture for reconstruction}
\end{figure}

\subsection{Application of Fourier Transform}
The Fourier transform of a 2D image refers to the process of transforming the image from the spatial domain to the frequency domain, allowing us to obtain information related to the frequency components of the image. The Fourier transform is used in image processing and analysis by converting the original image to the frequency domain, performing necessary operations, and then restoring it back to the original domain through inverse transformation.\\
First, the given 2D square image (256, 256) is represented in the spatial domain using $(x,y)$ coordinates. To perform the Fourier transform on this image, we use Eq. (2): $F(u,v)$ is a complex number representing the transformed result in the frequency domain, and $(u,v)$ represents the coordinates in the frequency domain. Eq. (2) implies multiplying the complex exponential function in the frequency domain and the image value at each position in the spatial domain, and then summing them up for all positions. This allows us to obtain frequency information of the image in the frequency domain.\\

\begin{equation}
F(u,v) = \sum_{x=0}^{N-1} \sum_{y=0}^{N-1} f(x,y) \cdot e^{-j2\pi \left(\frac{ux+vy}{N}\right)}
 \label{eq2}
\end{equation}


The inverse Fourier transform is represented by Eq. (3). Importantly, the property of the Fourier transform is that when the original image is transformed and then inverse transformed, it is restored to the original image. This represents the relationship between the Fourier transform and the inverse Fourier transform.

\begin{equation}
f(x,y) = \frac{1}{N^2} \sum_{u=0}^{N-1} \sum_{v=0}^{N-1} F(u,v) \cdot e^{j2\pi \left(\frac{ux+vy}{N}\right)}
\label{eq3}
\end{equation}


In this paper, multiple reconstructed results for a "Normal Reconstructed Template" representing normal data are defined as $T(x_i)$. To determine the presence of defects, both the target image to be evaluated and the Normal Reconstructed Template are Fourier transformed to convert them into the frequency domain. In the frequency domain, the part with frequency 0 is placed at the center, and as the frequency increases, it is shifted towards the edges of the frequency area through a shift process.\\
Next, a "Fourier Mask" is defined to remove the low-frequency components, as shown in Fig. 3. For this purpose, a square mask with a side length of $\tau$ centered at the origin is created. This mask is used to perform pixel-wise operations with the Fourier transformed result, effectively removing the low-frequency components. This process retains only the high-frequency region where defects exist, while eliminating the background and unwanted components. Finally, an inverse transformation is applied to convert the result back from the frequency domain to the spatial domain. As a result, only the defective regions of the Texture image are preserved in the spatial domain.


\begin{figure}[htb!]
    \centering
    \includegraphics[width=16cm]{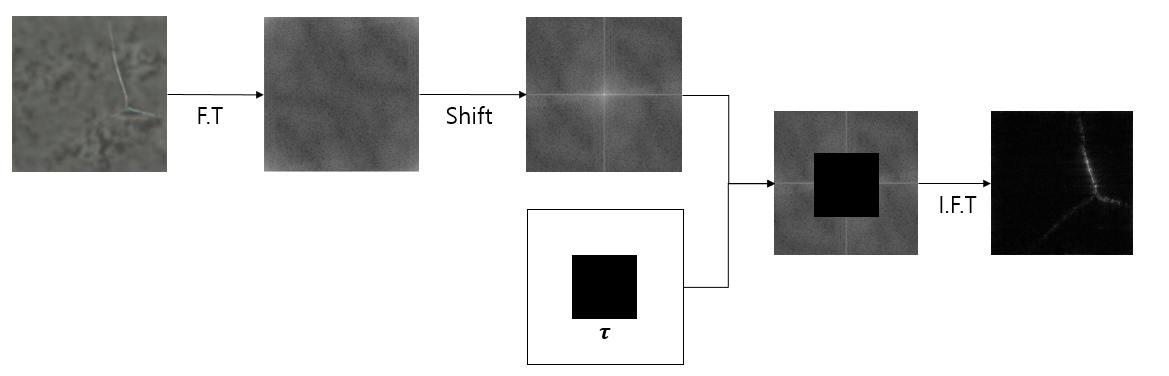}
    \caption{Fourier Mask and Pixel-wise}
\end{figure}

\subsection{Difference Images and Thresholding}
The two generated images ($F(R(x))$, $F(T(x_i))$) obtained through the given process represent the result of removing the low-frequency band and retaining only the high-frequency band. Therefore, the Normal Reconstructed Template retains only fine high-frequency details while removing the rest. If we perform the same process using a normal image as input, the difference with the Normal Reconstructed Template will be very small. However, when an image with defects is used as input, the defective parts are not completely eliminated through the process, and some noise from the background may remain. By subtracting the two generated images, the resulting difference image will have non-zero values in the defective regions and values close to zero in the remaining areas. This difference image can be used to create a final map for defect detection. Finally, by applying a threshold value $th$ to the generated map, we can generate final maps that allow us to determine the presence of defects. If the values exceed the threshold, they are set to one. Otherwise, they are set to zero. This binary map indicates the presence of defects where it is set to one and absence where it is set to zero. By calculating the total count of pixels in the generated map, we can derive scores for each image. Since the score range can vary significantly, we normalize the scores based on the scores of all the images and calculate an appropriate defect score.\\
Therefore, the resulting binary image obtained through this process will have one in the locations of defects and zero in the unaffected areas. This enables defect detection, and by calculating scores for the entire image and normalizing them, we can determine the normalized defect score. Fig. 4 represents the process of calculating the defect score.


\begin{figure}[htb!]
    \centering
    \includegraphics[width=16cm]{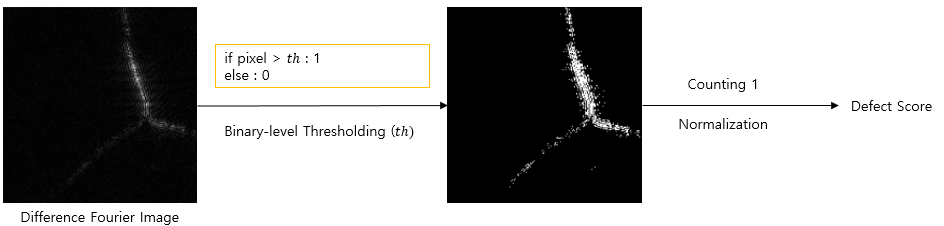}
    \caption{Process of calculating the defect score}
\end{figure}


\section{Experimental Results}
\subsection{Datasets}

In this paper, we focus on Texture Defect Detection using the MV-Tec AD[20]. This datasets consist of 5 textures and 10 objects, but we only evaluate the performance using 5 textures. However, the available data is insufficient for both training and testing. Therefore, data augmentation is performed in this paper. Tab. 1 represents the final composition of the datasets in terms of the number of samples for each category. Since we only use normal data for training, the majority of the existing data is used as training data. To balance the number of samples between normal and defect data, augmentation techniques are applied to the normal data. This process increases the diversity of the training data and ensures an adequate number of defect samples, ultimately improving the performance of the model.


\begin{table}[htbp]
 \caption{Detailed datasets with data augmentation}
  \centering
  \begin{tabular}{lrrrr}
    \toprule
    Category  &Train(Normal) &Test(Normal) &Test(Defect)\\
     \midrule
    Carpet &280 &84 &89\\
    Grid &264 &63 &57\\
    Leather &245 &96 &92\\
    Tile &230 &99 &84\\
    Wood &247 &57 &60\\
   
    \bottomrule
  \end{tabular}
  \label{tab:table}
\end{table}

\subsection{Training Details}

For Texture Defect Detection in this study, the following training approach is utilized. The input normal image data is normalized within the range of 0 and 1, and data augmentation is performed using ImageDataGenerator to generate diverse forms of training data. Data augmentation techniques such as shearing (20\%), zooming (20\%), and vertical and horizontal flipping are applied. During the training process, the Adam optimizer is employed with an initial learning rate set to 1e-4. The training is conducted for 500 epochs on the entire datasets, with a batch size of 16. In the loss function, the hyperparameter $\lambda_{L2}$ is set to 100 for L2 loss, and $\lambda_{L1}$ is set to 1 for L1 loss, resulting in a combination of simple loss functions.


\subsection{Performance Evaluation and Ablation Study}

The proposed network in this study is trained only on normal data, resulting in the generation of images with different distributions when dealing with defect images. Fig. 5 illustrates the inferred reconstructed images from examples of defective data for each category. All the data in the figure contains defects. The first row shows the original images with defects, while the second row represents the reconstructed images. The original images exhibit patterns even in the background, indicating prominent features. However, the defective regions possess even more pronounced features. Therefore, by removing the noise in the background, the defective regions can be highlighted more effectively. The overall reconstructed results appear blurred, with a significant reduction in noise in the background except for the grid. Although the defective regions also become blurry, the removal of background patterns makes it easier to extract the defects. The grid exhibits a consistent pattern and closely resembles the original, except for the areas with defects where differences can be observed.

\begin{figure}[htb!]
    \centering
    \includegraphics[width=16cm]{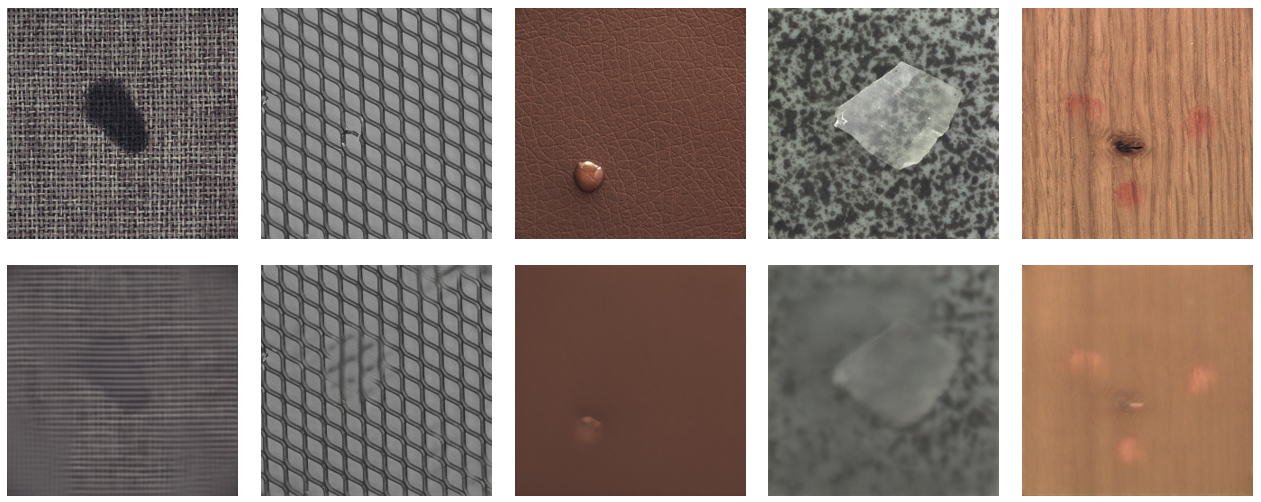}
    \caption{Reconstructed image with defect}
\end{figure}

The reason for using Normal Reconstructed Templates is to overcome the limitations of the network and improve performance by assessing the differences between the closest reconstructed image and the original image, even for normal data. In reality, even normal data cannot be perfectly reconstructed to match the original. Hence, the approach involves extracting the normal regions by utilizing the differences between the reconstructed image and the original image as closely as possible. Normal Reconstructed Template represents the restoration results of normal data and possess various forms and patterns. These templates are used to generate difference Fourier images, which are then employed to detect defective regions. As the network is trained solely on normal data, processing data with defects will result in reconstructed images with slightly different distributions. Thus, the approach involves calculating the differences between the normal reconstructed templates and the defective regions to extract the defects.\\
This approach allows for the precise detection of defects by distinguishing between normal and defective regions. By removing the normal regions based on the differences with the reconstructed templates, the remaining areas are composed of the defective regions, making the defects more distinct and enabling more accurate detection. Therefore, in this study, the proposal of utilizing normal reconstructed templates aims to overcome the limitations of the network and enhance performance. By removing the normal regions and emphasizing the defective regions, more accurate defect detection can be achieved. The reconstructed images from normal data exhibit a variety of normal restoration templates. Therefore, it is crucial to select the most suitable template for evaluating the reconstructed results on test normal images. Fig. 6 presents the selection of appropriate templates for each category based on experimental results. These templates can be utilized to generate difference Fourier images and improve performance consistently across all data. Additionally, when generating difference Fourier images, the 10-pixel edge is excluded from evaluation. This is because the edge exhibits a different distribution compared to the original due to padding, increasing the likelihood of misclassification.\\

\begin{figure}[htb!]
    \centering
    \includegraphics[width=16cm]{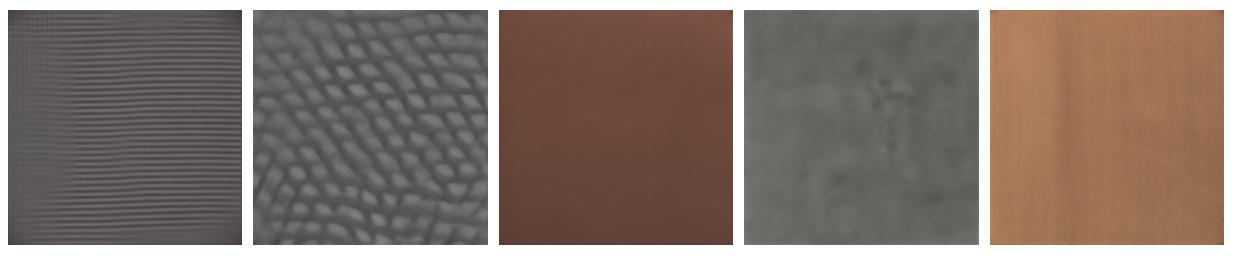}
    \caption{Normal Reconstructed Template}
\end{figure}

As mentioned in Chapter 3, we need to find the most suitable template by combining the length of the Fourier mask, denoted as $\tau$, and the binary-level thresholding represented by $th$. To evaluate the performance, we use the Area Under the Curve (AUC) as the evaluation metric. AUC is a common metric used to assess the performance of classification models, ranging from 0 to 1, where a value closer to 1 indicates better performance. Since each category has different characteristics, they have different parameter values, and we explore various combinations of these parameters. Therefore, we use AUC to find the optimal parameter combination for each category and select the combination with the highest AUC value. Based on this, we inferred the AUC values for various parameter combinations and presented the results in Table 2. The parameter combinations that showed the highest AUC for each category are as follows ($\tau$, $th$): Carpet: (40, 13), Grid: (43, 16), Leather: (3, 4), Tile: (40, 2), Wood: (2, 9).\\
The overall average AUC is 93.1\%. This indicates that our proposed simple method achieves performance similar to state-of-the-art approaches. These results demonstrate that our method effectively detects defects despite its simplicity. However, the Carpet category recorded a relatively lower AUC compared to other categories. This can be attributed to the relatively smaller difference between the defect regions and the background in the Carpet category compared to other categories. When the difference between defect regions and the background is small, removing noise from the background may also result in the removal of defect regions, making accurate detection more challenging. Therefore, the Carpet category may require different parameter values and additional adjustments.\\
In conclusion, we have shown that a simple method can achieve high performance. Additionally, since optimal parameter combinations may vary for each category, it is important to adjust parameter values accordingly. Thus, our method offers both flexibility and simplicity, making it applicable to various defect detection problems.\\
Fig. 7 shows a partial process of generating the final decision map for each category. The first column represents the reconstructed image. The second and third columns show frequency domain images of the input image and normal reconstructed template, respectively. The fourth column illustrates the result of binary-level thresholding applied to the difference Fourier image. The white areas indicate defective regions, and it can be observed that the actual defective areas are well preserved.


\begin{table}[htbp]
\caption{Performance analysis (row: Fourier mask length $\tau$, column: threshold value $th$) }
\centering
\begin{tabular}{c|cccccc}
\hline
Carpet & 9 & 10 & 11 & 12 & 13 & 14 \\
\hline
37 & 0.831 & 0.790 & 0.753 & 0.720 & 0.687 & 0.679 \\
38 & 0.771 & 0.751 & 0.733 & 0.719 & 0.705 & 0.697 \\
39 & 0.798 & 0.779 & 0.767 & 0.761 & 0.764 & 0.762 \\
40 & 0.852 & 0.848 & 0.844 & 0.852 & \textbf{0.854} & 0.829 \\
41 & 0.836 & 0.825 & 0.796 & 0.796 & 0.728 & 0.663 \\
42 & 0.796 & 0.768 & 0.752 & 0.715 & 0.645 & 0.566 \\
\hline\hline

Grid & 15 & 16 & 17 & 18 & 19 & 20 \\
\hline
42 & 0.917 & 0.906 & 0.929 & 0.926 & 0.923 & 0.911 \\
43 & 0.913 & \textbf{0.934} & 0.927 & 0.925 & 0.920 & 0.912 \\
44 & 0.920 & 0.923 & 0.926 & 0.926 & 0.918 & 0.884 \\
45 & 0.924 & 0.927 & 0.918 & 0.914 & 0.893 & 0.875 \\
46 & 0.916 & 0.923 & 0.918 & 0.915 & 0.901 & 0.884 \\
47 & 0.913 & 0.912 & 0.901 & 0.899 & 0.896 & 0.873 \\
\hline\hline

Leather & 1 & 2 & 3 & 4 & 5 & 6 \\
\hline
1 & 0.781 & 0.728 & 0.736 & 0.843 & 0.926 & 0.961 \\
2 & 0.746 & 0.792 & 0.882 & 0.935 & 0.926 & 0.868 \\
3 & 0.752 & 0.865 & 0.946 & \textbf{0.974} & 0.916 & 0.871 \\
4 & 0.633 & 0.818 & 0.908 & 0.931 & 0.900 & 0.855 \\
5 & 0.562 & 0.794 & 0.916 & 0.906 & 0.897 & 0.853 \\
6 & 0.513 & 0.768 & 0.891 & 0.916 & 0.890 & 0.835 \\
\hline\hline

Tile & 1 & 2 & 3 & 4 & 5 & 6 \\
\hline
37 & 0.729 & 0.907 & 0.752 & 0.95 & 0.643 & 0.607 \\
38 & 0.729 & 0.893 & 0.744 & 0.687 & 0.643 & 0.607 \\
39 & 0.746 & 0.920 & 0.749 & 0.687 & 0.637 & 0.607 \\
40 & 0.747 & \textbf{0.921} & 0.949 & 0.681 & 0.625 & 0.601 \\
41 & 0.761 & 0.919 & 0.739 & 0.675 & 0.619 & 0.601 \\
42 & 0.762 & 0.892 & 0.737 & 0.675 & 0.613 & 0.601 \\
\hline\hline

Wood & 7 & 8 & 9 & 10 & 11 & 12 \\
\hline
1 & 0.834 & 0.835 & 0.860 & 0.898 & 0.913 & 0.919 \\
2 & 0.899 & 0.925 & \textbf{0.971} & 0.961 & 0.934 & 0.919 \\
3 & 0.916 & 0.946 & 0.938 & 0.938 & 0.928 & 0.901 \\
4 & 0.923 & 0.921 & 0.926 & 0.918 & 0.917 & 0.911 \\
5 & 0.929 & 0.911 & 0.909 & 0.905 & 0.908 & 0.900 \\
6 & 0.895 & 0.900 & 0.897 & 0.906 & 0.897 & 0.884 \\
\hline

\end{tabular}
\label{table:carpet-auc}
\end{table}

\begin{figure}[htb!]
    \centering
    \includegraphics[width=16cm]{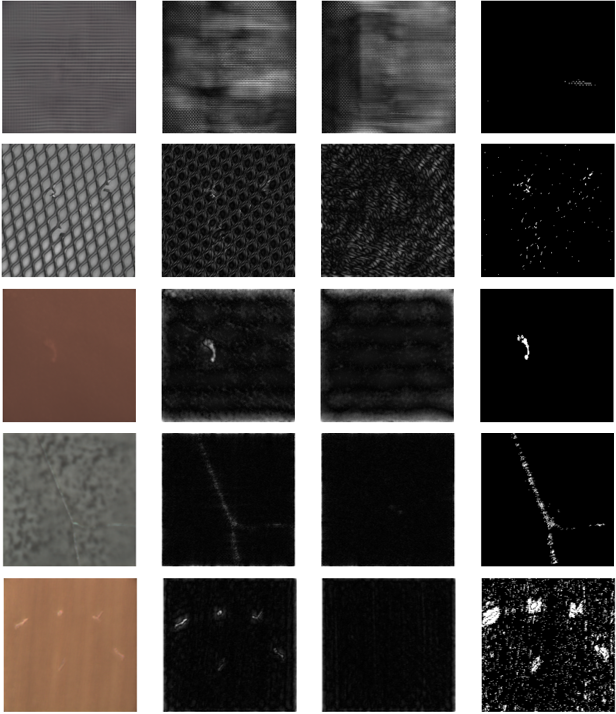}
    \caption{Process for locating defect detection (1st column: reconstruction, 2nd-3rd columns: Fourier trans-form results, 4th column: binary-level thresholding)}
\end{figure}


Tab. 3 presents the results of evaluating various anomaly detection algorithms on different categories. The performance of each algorithm is measured by the AUC (Area Under the Curve) value, where a value of 1 indicates the presence of defects and 0 indicates their absence.\\
Analyzing the results, we can observe that the algorithm DAAD[19] achieved the highest AUC value of 0.866 for the Carpet category. This indicates that DAAD[19] demonstrates effective performance in anomaly detection for Carpet data. Additionally, for the Grid and Wood category, DAAD[19] also shows a high AUC value of 0.957 and 0.982, indicating its superior performance in anomaly detection. For the Leather category, the proposed method achieves the highest AUC value of 0.974. This suggests that proposed method have strong anomaly detection performance for Leather data, respectively. For the Tile category, proposed methods obtain the highest AUC value of 0.921. This indicates that the proposed method demonstrates excellent anomaly detection capability for Tile data.\\
Overall, AnoGAN[1] shows relatively lower performance across all categories, while GANomaly[3] and Skip-GANomaly[4] exhibit highly irregular results compared to other categories. DAAD[19] demonstrates high performance for specific categories, but the proposed method shows excellent performance in all categories except Carpet. Notably, the proposed method stands out as it does not heavily rely on deep learning networks compared to previous studies, and it exhibits the smallest category-wise variation with the highest average AUC of 93.1\%. The significant difference observed in the Leather category can be attributed to the simplicity of the background, making noise removal easier.


\begin{table}[htbp]
\caption{: Performance analysis (row: Fourier mask length $\tau$, column: threshold value $th$) }
\centering
\begin{tabular}{l|ccccccc|}
\hline
Methods & AnoGAN[1] & memAE[5] & OCGAN[6] & GANomaly[3] & Skip-GANomaly[4] & DAAD[19] & Ours \\
\hline
Carpet & 0.337 & 0.386 & 0.348 & 0.699 & 0.795 & \textbf{0.866} & \underline{0.854} \\
Grid & 0.871 & 0.805 & 0.855 & 0.708 & 0.657 & \textbf{0.957} & \underline{0.934} \\
Leather & 0.451 & 0.423 & 0.624 & 0.842 & \underline{0.908} & 0.862 & \textbf{0.974} \\
Tile & 0.401 & 0.718 & 0.806 & 0.794 & 0.850 & 0.882 & \textbf{0.921} \\
Wood & 0.567 & 0.954 & 0.959 & 0.834 & 0.919 & \textbf{0.982} & \underline{0.971} \\
\hline
Average & 0.525 & 0.657 & 0.718 & 0.775 & 0.826 & \underline{0.910} & \textbf{0.931} \\
\hline
\end{tabular}
\label{table:method-comparison}
\end{table}


The proposed method in this paper has two main aspects: using a simple autoencoder for re-construction and combining it with Fourier transform to separate and remove defective features. Therefore, the performance is evaluated based on the usage of reconstruction and Fourier trans-form. Without reconstruction, performing only Fourier transform makes it extremely challenging to separate defective features in the frequency domain, resulting in a slightly higher AUC of 53.1\% overall compared to random results. When only reconstruction is used along with Fourier transform, it yields very low performance with an average AUC of 63.5\% across all categories except for Tile. This indicates that using only reconstruction to preserve defective regions is difficult for other categories, except for Tile, where the reconstructed results alone can preserve defective regions. On the other hand, the proposed method that combines both reconstruction and Fourier transform achieves the highest performance with an AUC of 93.1\%. Although there is a 2.6\% AUC decrease for Tile, there is an overall improvement of up to 46.6\%, making the proposed method valid.

\begin{table}[htbp]
\caption{: Performance analysis (row: Fourier mask length $\tau$, column: threshold value $th$) }
\centering
\begin{tabular}{l|ccc}
\hline
 Reconstruction &X &O &O\\
 Fourier Transform &O &X &O\\
\hline
Carpet & 0.506 & 0.697 & \textbf{0.854} \\
Grid & 0.634 & 0.508 & \textbf{0.934} \\
Leather & 0.505 & 0.517 & \textbf{0.974} \\
Tile & 0.505 & \textbf{0.947} & 0.921 \\
Wood & 0.503 & 0.505 &\textbf{0.971} \\
\hline
Average & 0.531 & 0.635 & \textbf{0.931} \\
\hline
\end{tabular}
\label{table:method-comparison}
\end{table}

\section{Conclusions}
This study proposes a method in the field of image analysis that combines a simple autoencoder and Fourier transform for texture defect detection. The experimental results showed that the pro-posed method achieved overall high performance, although there were cases where it did not show effectiveness for some data. Therefore, for future research, it is necessary to enhance the performance of the proposed method by utilizing deep learning networks to improve noise removal. By developing and applying more sophisticated denoising techniques through deep learning networks, the defect detection performance can be further improved. Additionally, future research directions include conducting comparative studies with other anomaly detection algorithms to validate the superiority of the proposed method and performing experiments on a wider range of datasets to evaluate its generalization performance.


\begin{thebibliography}{1}

\bibitem{andrews2016anogan}
Andrews, W., Benitez, V., He, Z., Khan, A., and Soyer, H.P.,
\newblock "AnoGAN: Unsupervised anomaly detection with generative adversarial networks to guide marker discovery," 
\newblock In \emph{European Conference on Computer Vision}, pp. 823-834, 2016.

\bibitem{schlegl2017unsupervised}
Schlegl, T., Seeböck, P., Waldstein, S.M., Schmidt-Erfurth, U., and Langs, G.,
\newblock "Unsupervised anomaly detection with generative adversarial networks to guide marker discovery," 
\newblock In \emph{International Conference on Information Processing in Medical Imaging}, pp. 146-157, 2017.

\bibitem{akcay2018ganomaly}
Akçay, S., Atapour-Abarghouei, A., and Breckon, T.P.,
\newblock "GANomaly: Semi-supervised anomaly detection via adversarial training," 
\newblock In \emph{Asian Conference on Computer Vision}, pp. 622-637, 2018.

\bibitem{akcay2018skipganomaly}
Akçay, S., Atapour-Abarghouei, A., and Breckon, T.P.,
\newblock "Skip-GANomaly: Skip connected and adversarially trained encoder-decoder anomaly detection," 
\newblock In \emph{Proceedings of the IEEE Conference on Computer Vision and Pattern Recognition Workshops}, pp. 39-42, 2018.

\bibitem{gong2019memorizing}
Gong, D., Liu, L., Le, V., Saha, B., Mansour, M.R., Venkatesh, S., and Van Den Hengel, A.,
\newblock "Memorizing normality to detect anomaly: Memory-augmented deep autoencoder for unsupervised anomaly detection," 
\newblock In \emph{Proceedings of the IEEE/CVF International Conference on Computer Vision}, pp. 1705-1714, 2019.

\bibitem{perera2019ocgan}
Perera, P., Nallapati, R., and Xiang, B.,
\newblock "OCGAN: One-class novelty detection using GANs with constrained latent representations," 
\newblock In \emph{Proceedings of the IEEE/CVF Conference on Computer Vision and Pattern Recognition}, pp. 2898-2906, 2019.

\bibitem{zhao2021plant}
Zhao, Y., Chen, Z., Gao, X., Song, W., Xiong, Q., Hu, J., and Zhang, Z.,
\newblock "Plant disease detection using generated leaves based on DoubleGAN," 
\newblock \emph{IEEE/ACM Transactions on Computational Biology and Bioinformatics}, Vol. 19, No. 3, pp. 1817-1826, 2021.

\bibitem{si2023chili}
Si, J., and Kim, S.,
\newblock "Chili Pepper Disease Diagnosis via Image Reconstruction Using GrabCut and Generative Adversarial Serial Autoencoder," 
\newblock \emph{arXiv preprint arXiv:2306.12057}, pp. 1-12, 2023.

\bibitem{si2021traffic}
Si, J., and Kim, S.,
\newblock "Traffic Accident Detection in First-Person Videos Based on Depth and Background Motion Estimation," 
\newblock \emph{Journal of Korean Institute of Information Technology (JKIIT)}, Vol. 19, No. 3, pp. 25-34, 2021.

\bibitem{tsai2018defect}
Tsai, D.M., and Huang, C.K.,
\newblock "Defect detection in electronic surfaces using template-based Fourier image reconstruction," 
\newblock \emph{IEEE Transactions on Components, Packaging and Manufacturing Technology}, Vol. 9, No. 1, pp. 163-172, 2018.

\bibitem{si2022surface}
Si, J., and Kim, S.,
\newblock "Surface Anomaly Detection of Wood Grain Image Using Fourier Transform: A Preliminary Study," 
\newblock In \emph{Proceedings of Korean Institute of Information Technology Conference}, pp. 86-87, 2022.

\bibitem{zavrtanik2021draem}
Zavrtanik, V., Kristan, M., and Skočaj, D.,
\newblock "Draem: A discriminatively trained reconstruction embedding for surface anomaly detection," 
\newblock In \emph{Proceedings of the IEEE/CVF International Conference on Computer Vision}, pp. 8330-8339, 2021.

\bibitem{liang2023omni}
Liang, Y., Zhang, J., Zhao, S., Wu, R., Liu, Y., and Pan, S.,
\newblock "Omni-frequency channel-selection representations for unsupervised anomaly detection," 
\newblock \emph{arXiv preprint arXiv:2203.00259}, pp. 1-14, 2023.

\bibitem{jang2023n-pad}
Jang, J., Hwang, E., and Park, S.H.,
\newblock "N-pad: Neighboring pixel-based industrial anomaly detection," 
\newblock In \emph{Proceedings of the IEEE/CVF Conference on Computer Vision and Pattern Recognition}, pp. 4364-4373, 2023.

\bibitem{si2023difference}
Si, J., and Kim, S.,
\newblock "Difference Image Alignment Technique of Reconstruction Method for Detecting Defects in Thermal Image of Solar Cells," 
\newblock \emph{Journal of Korean Institute of Information Technology (JKIIT)}, Vol. 21, No. 5, pp. 11-19, 2023.

\bibitem{tsai2022multi}
Tsai, C.C., Wu, T.H., and Lai, S.H.,
\newblock "Multi-scale patch-based representation learning for image anomaly detection and segmentation," 
\newblock In \emph{Proceedings of the IEEE/CVF Winter Conference on Applications of Computer Vision}, pp. 3992-4000, 2022.

\bibitem{liu2022reconstruction}
Liu, T., Li, B., Zhao, Z., Du, X., Jiang, B., and Geng, L.,
\newblock "Reconstruction from edge image combined with color and gradient difference for industrial surface anomaly detection," 
\newblock \emph{arXiv preprint arXiv:2210.14485}, pp. 1-11, 2022.

\bibitem{shi2021unsupervised}
Shi, Y., Yang, J., and Qi, Z.,
\newblock "Unsupervised anomaly segmentation via deep feature reconstruction," 
\newblock \emph{Neurocomputing}, Vol. 424, pp. 9-22, 2021.

\bibitem{jinlei2021divide}
Jinlei, H., Yingying, Z., Qiaoyong, Z., Di, X., Shiliang, P., and Hong, Z.,
\newblock "Divide-and-assemble: Learning block-wise memory for unsupervised anomaly detection," 
\newblock In \emph{Proceedings of the IEEE/CVF International Conference on Computer Vision (ICCV)}, pp. 8791-8800, 2021.

\bibitem{bergmann2019mvtec}
Bergmann, P., Fauser, M., Sattlegger, D., and Steger, C.,
\newblock "MVTec AD--A comprehensive real-world dataset for unsupervised anomaly detection," 
\newblock In \emph{Proceedings of the IEEE/CVF Conference on Computer Vision and Pattern Recognition}, pp. 9592-9600, 2019.

\end{thebibliography}


\end{document}

\end{document}